\documentclass[11pt]{article}

\usepackage[preprint]{acl}

\usepackage{times}
\usepackage{latexsym}

\usepackage[T1]{fontenc}

\usepackage[utf8]{inputenc}

\usepackage{microtype}

\usepackage{inconsolata}

\usepackage{graphicx}
\usepackage{amsmath, amsfonts}
\usepackage{float}
\usepackage{amsmath, amsfonts}
\usepackage{booktabs}
\usepackage{multirow}
\usepackage{svg}
\usepackage{subcaption}

\usepackage[dvipsnames]{xcolor}

\title{SALSA: Speech Aware LLM Adaptation via Learned Steering Activation Vectors}

\author{
  \textbf{Yekaterina Yegorova\textsuperscript{1}},
  \textbf{Argyrios Gerogiannis\textsuperscript{1}},
  \textbf{Haolong Zheng\textsuperscript{1}},
\\
  \textbf{Julia Hockenmaier\textsuperscript{1}},
  \textbf{Chang D. Yoo\textsuperscript{2}},
  \textbf{Mark A. Hasegawa-Johnson\textsuperscript{1}}
\\
\\
  \textsuperscript{1}University of Illinois Urbana-Champaign,
  \textsuperscript{2}Korea Advanced Institute of Science and Technology
\\
  \texttt{\{yay2, ag91, haolong2, juliahmr, jhasegaw\}@illinois.edu}
\\
  \texttt{cd\_yoo@kaist.ac.kr}
}

\begin{document}
\maketitle
\begin{abstract}

Speech-aware large language models often generalize poorly to out-of-domain settings. We propose SALSA (Speech-Aware LLM Adaptation via Learned Steering Activations), a lightweight adaptation method that learns layer-wise steering vectors. Unlike commonly used steering approaches that rely on contrastive activation differences, SALSA directly optimizes steering vectors using a supervised objective. Across children’s speech, multilingual speech, and Mandarin-English code-switching benchmarks, SALSA substantially improves performance over zero-shot inference and speech in-context learning baselines, achieving up to 46.8\% relative improvements over zero-shot. Analysis further demonstrates that steering the encoder, particularly the later layers, is more effective than steering the LLM backbone. These findings suggest that steering improves downstream ASR performance by adapting higher-level acoustic and phonetic representations to better align with the pretrained language model representation space, rather than by modifying the decoder itself.

\end{abstract}

\section{Introduction}

Speech-aware large language models (SALLMs) have emerged as a powerful paradigm for speech processing, leveraging linguistic knowledge encoded in pretrained large language models (LLMs) to advance performance across a broad range of speech tasks \citep{peng2025survey}. Despite their strong performance on high-resource languages, they generalize poorly to out-of-domain scenarios, and this occurs even when the underlying model components have already been exposed to the target languages or tasks during pretraining. This suggests that adaptation in SALLMs is not solely a problem of acquiring new linguistic knowledge, but also one of aligning pretrained acoustic representations with the language model decoding space for downstream ASR tasks. Fine-tuning can address these gaps, but is computationally expensive and data-intensive. Parameter-efficient alternatives such as LoRA \citep{hu_lora_2021} reduce this cost but still require gradient updates to model weights. In-context learning (ICL) \citep{NEURIPS2020_1457c0d6} is training-free, but is constrained by the high acoustic variability of speech. Unlike text, speech encodes speaker identity and phonetic properties alongside semantics, making it difficult to identify informative demonstration examples \citep{zheng2025ticl}.

A complementary method is activation steering: directly perturbing a model's internal representations at inference time to shift its behavior toward a target domain, without modifying its weights. Steering has been shown to be effective in language models \cite{im2025unified} and has recently started to be explored for speech \cite{sun2026activation, feng2025steer}. Existing approaches for the speech modality rely on contrastive activation differences, where steering vectors are derived from paired contrastive examples. For automatic speech recognition (ASR), natural paired examples are difficult to obtain, unlike for semantic tasks. ASR requires a direct correspondence between the input audio and output transcript, and the high acoustic variability of speech makes contrastive pairs noisy and hard to construct. This limits the applicability of existing steering methods to exactly the low-resource, out-of-domain settings where efficient adaptation is most needed.

In this paper, we propose SALSA (\textbf{S}peech-aware LLM \textbf{A}daptation via \textbf{L}earned \textbf{S}teering \textbf{A}ctivations), a lightweight adaptation method that directly learns layer-wise steering vectors for speech-language models without requiring paired contrastive examples. By training steering vectors, SALSA can learn around acoustic variability and adapt intermediate encoder representations to better align with the pretrained language model representation space. Our contributions are as follows:

\begin{itemize}
    \item We propose SALSA, a lightweight steering-based adaptation method that learns layer-wise encoder steering vectors for pretrained SALLMs while keeping all backbone parameters frozen.

    \item  We demonstrate that simple learned steering vectors are highly effective for out-of-domain ASR settings, including children’s speech, multilingual speech, and code-switched speech, substantially improving over zero-shot and speech in-context learning baselines without requiring paired contrastive examples.

    \item We analyze steering behavior across training scale, encoder depth, and intervention location, showing that adaptation is most effective in higher-level encoder representations, while steering the language model backbone provides substantially smaller gains than encoder steering.

\end{itemize}

\section{Background}

SALLMs, such as SALMONN \citep{tang2024salmonn}, WavLLM \citep{hu2024wavllm}, the Qwen-Audio family \citep{chu2023qwenaudioadvancinguniversalaudio, Qwen2-Audio}, and Granite-speech \citep{saon2025granite}, are typically constructed by coupling a pretrained speech encoder with an LLM backbone, where only a lightweight projection layer is trained to bridge the two modalities. These models support a broad range of tasks, from traditional speech processing tasks such as ASR to more open-ended tasks, including conversational speech and question answering. However, relative to speech-native models such as Whisper \citep{radford2023robust}, SALLMs are generally trained on considerably less speech data and tend to focus on a limited set of predominantly European languages. As a result, SALLMs are prone to hallucination and generalize poorly to out-of-domain data.

\subsection{Adaptation Methods for SALLMs}

Adaptation through full fine-tuning or parameter-efficient methods such as LoRA \citep{hu_lora_2021} can address out-of-domain limitations, but such approaches require modifying model parameters and may be impractical in low-resource or computationally-constrained settings. More recently, reinforcement learning (RL)-based post-training methods have also been explored for improving speech-language models, although these approaches similarly require iterative model updates and large-scale optimization \citep{guo2025deepseek, rafailov2023direct, elmakies2026advancing}. A lighter-weight alternative is speech in-context learning (SICL) \citep{SICL}, which was originally proposed for Whisper and has since been extended to SALLMs \citep{omnilingual2025omnilingual, abouelenin2025phi, roll2025context}. \citet{zheng2026ticl} further demonstrates that selecting semantically similar in-context examples substantially improves performance across a range of challenging conditions, including children's speech, accented speech, and multilingual settings. However, unlike text, speech contains substantial variability in speaker identity, prosody, and acoustic conditions, making it difficult to retrieve informative demonstrations across out-of-domain settings.

\subsection{Steering}
\label{sec:steering}

Representation steering methods aim to control model behavior by intervening directly in the hidden representations of a neural network during inference. Given a model with hidden representation $h_l \in \mathbb{R}^d$ at layer $l$, steering methods modify the forward pass by applying an intervention function
\[
\tilde{h}_l = f(h_l, v, \alpha),
\]
where $v$ denotes a steering direction or transformation and $\alpha$ controls the intervention strength. The modified representation $\tilde{h}_l$ is then propagated through the remainder of the network to influence model outputs.

Many steering methods use additive interventions of the form
\[
\tilde{h}_l = h_l + \alpha v,
\]
where $v$ corresponds to a direction in representation space associated with a desired behavior or attribute. Early work demonstrated that meaningful behavioral directions can often be extracted from contrastive examples. \citep{rimsky2024steering} computes steering vectors from differences between hidden activations of paired prompts exhibiting opposite behaviors. Given activations $h_l^{+}$ and $h_l^{-}$ corresponding to positive and negative examples, respectively, a steering vector can be computed as
\[
v = \frac{1}{N}\sum_{i=1}^{N} \left(h_{l,i}^{+} - h_{l,i}^{-}\right).
\]

This contrastive approach has been extended through PCA-based steering methods \citep{liu2024incontext, zou2023representation}, which estimate steering vectors from the principal components of contrastive activation differences. By identifying directions that explain the largest variance across contrastive pairs, PCA-based methods aim to isolate more robust and semantically meaningful steering directions. Approaches like Inference-Time Intervention \citep{li2023inference}, representation extraction methods \citep{subramani-etal-2022-extracting}, personalized steering techniques \citep{cao2024personalized}, and Representation Finetuning \citep{wu2024reft} learn steering transformations directly from data. Rather than using fixed additive vectors, these methods parameterize interventions as trainable functions,
\[
\tilde{h}_l = h_l + g_{\theta}(h_l),
\]
where $g_{\theta}$ may consist of lightweight modules trained to modify intermediate representations. Compared to the directly calculated steering vectors, learned interventions can capture more complex transformations.

While most prior steering work has focused on language models, steering has recently begun to expand into multimodal \citep{sivakumar2025steervlm, peng2024live, parekh2026learning} and speech settings. In speech-language models, activation steering has been explored to improve modality alignment across speech tasks \citep{feng2025steer} and to enhance robustness in out-of-distribution scenarios \citep{sun2026activation}. In these settings, steering interventions are typically applied to encoder representations prior to projection into the language model, with the goal of better aligning acoustic representations with the LLM token space.

\section{Steering Vector Training}

Prior steering methods commonly construct steering vectors from contrastive activation differences between paired examples, as demonstrated in Section \ref{sec:steering}. This paradigm has also recently been explored in speech-language models, where \citet{sun2026activation} construct steering directions from contrastive activation differences to improve ASR on accented speech. However, this formulation is poorly suited to speech tasks. Paired corpora are scarce in speech, particularly in out-of-domain settings, and the acoustic variability inherent to speech makes clean contrastive signals difficult to extract. Furthermore, unlike many semantic language generation tasks, ASR requires preserving a precise correspondence between acoustic inputs and textual outputs, leaving limited tolerance for semantic shifts that contrastive steering may produce.

Instead of extracting steering directions from activation differences, we directly optimize steering vectors using a supervised task objective. This removes the need for paired contrastive examples and learns interventions end-to-end in the representation space.

Let $f_\theta$ denote a frozen speech-language model with hidden representation $\mathbf{h}_l \in \mathbb{R}^d$ at layer $l$. We learn a set of layer-wise steering vectors $\mathcal{V} = \{\mathbf{v}_l\}_{l=1}^{L} $
where $\mathbf{v}_l \in \mathbb{R}^d$ and $L$ is the number of steered layers in the speech encoder. Following prior additive steering approaches, intermediate representations are modified via
\[
\tilde{\mathbf{h}}_l = \mathbf{h}_l + \mathbf{v}_l.
\]
During training, all parameters $\theta$ remain frozen and only the steering vectors $\mathcal{V}$ are optimized.

\paragraph{Steering Mechanism.}

To stabilize steering, we apply a norm-preserving update at each steered layer:
\begin{equation}
\tilde{\mathbf{h}}_l
=
\frac{\mathbf{h}_l + \mathbf{v}_l}
{\|\mathbf{h}_l + \mathbf{v}_l\|}
\cdot
\|\mathbf{h}_l\|.
\label{eq:steering}
\end{equation}
This renormalization preserves the original activation magnitude while modifying only the representation direction.

\paragraph{Training Objective.}

Let $p_{\theta,\mathcal{V}}$ denote the next-token distribution induced by the frozen model under steering intervention $\mathcal{V}$. Given an audio input $x$ and reference transcription $y^\star$, we optimize the steering vectors using the autoregressive cross-entropy objective
\begin{equation*}
\mathcal{L}(\mathcal{V})
=
-
\mathbb{E}_{(x,y^\star)}
\left[
\sum_{t=1}^{|y^\star|}
\log
p_{\theta,\mathcal{V}}
\left(
y_t^\star
\mid
x,
y^\star_{<t}
\right)
\right].
\end{equation*}
Since all backbone parameters remain frozen, optimization occurs entirely through representation-level interventions.

\paragraph{Optimization Details.}

All configurations share the same training setup. We optimize steering vectors using AdamW \citep{loshchilov2018decoupled} with learning rate  $\eta \in \{10^{-4}, 5{\times}10^{-4}\}$, batch size of 1, and gradient clipping with maximum norm $1.0$. Training proceeds for up to 20 epochs with early stopping (patience 3) based on validation WER.

\paragraph{Inference.} At inference time, the learned steering vectors $\mathcal{V}$ are injected into the selected layers using Eq.~\ref{eq:steering}. Encoder-only configurations apply steering to all audio encoder hidden states, while decoder-only configurations apply steering only to the audio-conditioned hidden representations within the language model backbone. Joint steering applies interventions to both modules simultaneously.

\section{Experiments}

\begin{table*}[!ht]
\vspace{-10pt}
\centering
\small
\label{tab:main-enc-combined}
\resizebox{\textwidth}{!}{%
\begin{tabular}{lllccccc}
\toprule
& & & \multicolumn{3}{c}{Children's Speech} & \multicolumn{2}{c}{SEAME}  \\
\cmidrule(lr){4-6} \cmidrule(lr){7-8}
Model & $n$ & System & MyST & OGI & RSR & dev-man & dev-sge \\
\midrule
\multirow{9}{*}{Granite-Speech-3.3-8B}
  & --  & zero-shot & 27.14  & 28.11 & 27.94 & 88.02 & 72.63 \\
\cmidrule(lr){2-8}
  & \multirow{3}{*}{500}
   & TICL      & $33.81_{\pm2.16}$   & $24.91_{\pm1.77}$   & $42.70_{\pm9.43}$    & $365.61_{\pm25.31}$   & $248.86_{\pm10.11}$\\
  & & SALSA     & $24.56_{\pm0.03}$ & $14.95_{\pm1.13}$ & $17.13_{\pm0.43}$ & $\mathbf{87.22_{\pm0.26}}$ & $\mathbf{73.49_{\pm1.17}}$ \\
\cmidrule(lr){2-8}
  & \multirow{3}{*}{2000}
  & TICL      & $31.78_{\pm5.34}$   & $17.35_{\pm0.20}$   & $38.85_{\pm3.85}$    & $336.97_{\pm33.02}$  & $224.39_{\pm11.39}$\\
  & & SALSA     & $\mathbf{24.10_{\pm0.07}}$ & $\mathbf{12.28_{\pm0.37}}$ & $\mathbf{14.46_{\pm0.39}}$ & $87.40_{\pm0.50}$ & $73.88_{\pm1.57}$ \\
\midrule
\multirow{9}{*}{Qwen2-Audio-7B-Instruct}
  & --  & zero-shot & 30.51 & 20.51 & 28.60 & 88.46 & 73.05 \\
\cmidrule(lr){2-8}
  & \multirow{2}{*}{500}
  & TICL      & $\mathbf{21.97_{\pm0.12}}$   & $16.14_{\pm0.66}$   & $30.29_{\pm6.48}$  & $188.75_{\pm18.93}$  & $174.46_{\pm17.76}$  \\
  & & SALSA     & $24.92_{\pm0.40}$ & $14.69_{\pm1.32}$ & $18.89_{\pm0.24}$ & $52.87_{\pm1.61}$  & $44.10_{\pm2.30}$ \\
\cmidrule(lr){2-8}
  & \multirow{2}{*}{2000}
  & TICL      & $22.09_{\pm0.23}$   & $14.40_{\pm9.54}$  & $28.19_{\pm2.26}$  & $173.59_{\pm4.36}$  & $159.12_{\pm11.32}$    \\
  & & SALSA     & $38.68_{\pm8.24}$ & $\mathbf{12.07_{\pm0.45}}$ & $\mathbf{15.84_{\pm0.33}}$ & $\mathbf{47.03_{\pm1.67}}$ & $\mathbf{40.12_{\pm1.42}}$  \\
\bottomrule
\end{tabular}%
}
\caption{WER (\%) ($\downarrow$) on Children's Speech datasets and MER (\%) ($\downarrow$) on SEAME for Qwen2-Audio-7B-Instruct and Granite-Speech-3.3-8B. Results are mean $\pm$ std across five random seeds.}
\label{tab:overview}
\end{table*}

\subsection{Datasets}
\paragraph{Children's Speech}

Children's speech differs systematically from adult speech in its acoustic and linguistic properties and is severely underrepresented in the training data of SALLMs. This makes children's speech a compelling scenario to evaluate whether steering can redirect models toward a structurally distinct and underrepresented population. The Redmond Speech Recall (RSR) Dataset \cite{redmond2019diagnostic, preza2026novel} is a corpus of children's speech (grades K-3) collected during a standardized screening task for developmental language disorder (DLD). The My Science Tutor (MyST) corpus \cite{pradhan2024my} consists of approximately 230 hours of conversational American English speech from elementary school children (grades 3-5) interacting with a virtual science tutor. The OGI Kids' Speech Corpus \cite{ogi} contains read and spontaneous speech from children ranging from grades K through 10.

\paragraph{Multilingual Speech}
CommonVoice 25.0 \citep{commonvoice:2020} is a massively crowd-sourced, multilingual speech corpus. We evaluate on two linguistically distinct subsets: Russian, a high-resource Slavic language, and Twi, a low-resource Kwa language spoken primarily in Ghana. Both Qwen2-Audio-7B-Instruct's LLM backbone and its speech encoder have been exposed to Russian separately during pre-training, whereas neither model has seen Twi. This pairing allows us to contrast adaptation to a language seen during pre-training against transfer to a language entirely absent from both models' training, letting us assess the extent to which pre-training language coverage can compensate for cross-modal representation gaps. SEAME \citep{lyu2010seame} is a conversational Mandarin-English code-switching corpus recorded from Singaporean and Malaysian speakers. Because both modules have seen Mandarin and English during training, SEAME tests a complementary capability: handling intra-utterance code-switching between two individually known languages, rather than transfer to an unseen one. 

Together, these corpora span three distinct multilingual conditions: pretraining coverage (Russian), complete absence of training exposure (Twi), and code-switching between known languages
(SEAME). This enables a structured analysis of multilingual representation within the models investigated.

\subsection{Models}
\paragraph{Qwen2-Audio-7B-Instruct}
Qwen2-Audio-7B-Instruct \citep{chu2024qwen2} is an open-weight SALLM that integrates a Whisper-based speech encoder with the pretrained QwenLM. The model supports two interaction modes: voice chat, where it responds directly to spoken input, and audio analysis, where the model conditions on both an audio signal and a text instruction. Qwen2-Audio-7B-Instruct supports 5 major European languages as well as Chinese, Cantonese, and Japanese. It is also instruction-tuned to follow natural language prompts about audio content.

\paragraph{Granite-Speech-3.3-8B}
Granite-Speech-3.3-8B \citep{saon2025granite} is a SALLM whose architecture contains a speech encoder built from 10 conformer blocks trained with Connectionist Temporal Classification loss \cite{graves2006connectionist} on ASR-focused data, a two-layer Q-Former \citep{pmlr-v202-li23q} projector that maps audio embeddings into the input space of the Granite-3.3-8B-Instruct LLM, with LoRA adapters (rank 64) applied to the query and value projection matrices. The model supports 5 major European languages, as well as English-to-Japanese and English-to-Mandarin translation.

\subsection{Baseline}

We compare against TICL \citep{zheng2026ticl}, an in-context learning adaptation method for speech-language models that conditions generation on the most semantically similar retrieved demonstrations from a candidate set. In experiments varying the amount of training data, the corresponding training split is additionally used as the retrieval candidate set for TICL. For Qwen2-Audio-7B-Instruct, retrieved demonstrations are prepended directly to the prompt following the standard TICL formulation. Granite-Speech-3.3-7B instead requires demonstrations to be formatted as dialogue history turns due to its input structure. Retrieved examples are therefore inserted as prior user-assistant interactions before the target query.

\subsection{Evaluation}

We evaluate transcription quality using Word Error Rate (WER) and Mixed Error Rate (MER). WER is the standard metric for automatic speech recognition (ASR), while MER is commonly used for Mandarin-English code-switched ASR.

Given the number of substitutions $S$, deletions $D$, insertions $I$, and reference tokens $N$, both metrics are computed as
\[
\mathrm{ErrorRate} = \frac{S + D + I}{N}.
\]

The metrics differ in how the reference and predicted transcriptions are tokenized prior to alignment. For WER, all text is tokenized at the word level. For MER, English speech is tokenized at the word level while Mandarin speech is tokenized at the character level, following standard evaluation protocols for Mandarin-English code-switched ASR. This mixed tokenization better reflects transcription quality across languages with different linguistic granularity.

All experiments use greedy autoregressive decoding, where the highest-probability token is selected at each generation step without beam search. WER and MER were computed using the jiwer package \citep{morris2004and} with standard text normalization and tokenization settings. Performance is reported as the mean and standard deviation across five independently sampled training subsets of size $n$, with random seeds 42-46 used for reproducibility.

\subsection{Results}

\begin{figure}[t]
    \centering
    \includesvg[width=\linewidth]{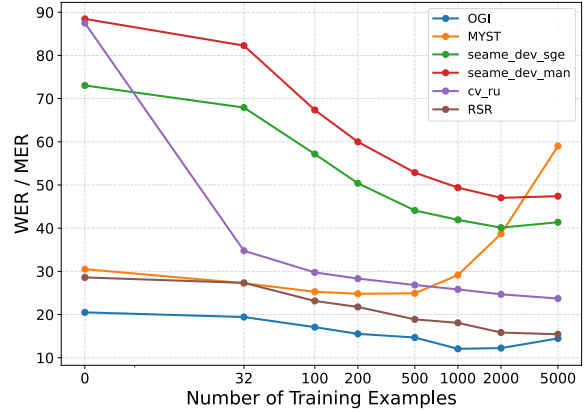}
    \caption{Scaling behavior of encoder steering on Qwen2-Audio-7B-Instruct across training set sizes. SALSA consistently improves performance on OGI, RSR, and SEAME, while MyST exhibits degradation at larger training sizes.}
    \label{fig:qwen_icv_scaling}
\end{figure}

Table \ref{tab:overview} reports performance on the Children's Speech datasets and SEAME using training sets of size $n \in \{500, 2000\}$ for both Qwen2-Audio-7B-Instruct and Granite-Speech-3.3-8B. Figure \ref{fig:qwen_icv_scaling} further shows scaling trends across a wider range of training sizes for Qwen2-Audio-7B-Instruct.

\paragraph{Qwen2-Audio-7B-Instruct.}
SALSA substantially improves performance over zero-shot prompting on OGI, RSR, and SEAME. On RSR, SALSA achieves a relative improvement of 44.6\% over zero-shot performance at $n=2000$, outperforming TICL by 43.8\%. Similarly, on OGI, SALSA improves over zero-shot by 42.3\%, while remaining competitive with TICL, differing by approximately 2 absolute WER points. These results suggest that encoder steering effectively adapts acoustic representations to better align with the decoder’s pretrained phonetic and linguistic representations.

In contrast, MyST exhibits different scaling behavior. While SALSA improves over zero-shot performance at smaller training sizes, performance degrades as additional steering data is introduced, as shown in Figure \ref{fig:qwen_icv_scaling}. At $n=1000$, TICL outperforms SALSA on MyST, suggesting that steering may be less effective for this dataset or more sensitive to overfitting. This may stem from the substantial intra- and inter-speaker variability in MyST, which may introduce inconsistent or noisy adaptation signals into the learned steering directions.

On SEAME, SALSA produces large gains on both evaluation splits, achieving up to a 46.8\% relative improvement on dev-man and 45.1\% on dev-sge compared to zero-shot prompting. Performance improvements begin to saturate beyond 2000 training samples. In contrast, TICL exhibits severe instability on SEAME, increasing MER by more than 100\% relative to the zero-shot baseline. This behavior may stem from interference between retrieved multilingual demonstrations and the model’s existing multilingual representations. In comparison, encoder steering appears to more effectively adapt the acoustic representations to the multilingual LLM backbone without disrupting the pretrained alignment between Mandarin and English representations.

Table~\ref{tab:cv} reports results on CommonVoice Russian and Twi. Across both datasets, SALSA consistently improves over zero-shot prompting, while TICL often provides little benefit or substantially degrades performance. On Russian, even 200 steering examples dramatically improve performance, suggesting that steering can effectively adapt the model for ASR in a language already seen during pretraining of both the encoder and LLM backbone. These results suggest that the primary challenge may not be language acquisition itself, but rather, aligning the encoder representations with the decoder’s transcription space for the downstream ASR task. On Twi, SALSA provides modest gains over zero-shot inference, although the limited amount of available training data makes it difficult to determine whether additional scaling would further improve performance.

\paragraph{Granite-Speech-3.3-8B.}
SALSA consistently improves over zero-shot inference on Granite-Speech-3.3-8B across the Children's Speech datasets. The largest gains are observed on OGI and RSR, where SALSA achieves relative WER reductions of 56.3\% and 48.2\%, respectively, at $n=2000$. SALSA also substantially outperforms TICL on these datasets, particularly on RSR, where TICL degrades performance relative to the zero-shot baseline.

Granite-Speech-3.3-8B additionally exhibits more stable scaling behavior than Qwen2-Audio-7B-Instruct as the number of steering examples increases from 500 to 2000. On MyST, SALSA achieves consistent improvements over zero-shot prompting, reducing WER from 27.14 to 24.10 at $n=2000$. This contrasts with the degradation observed for Qwen2-Audio-7B-Instruct on the same dataset and may suggest that Granite-Speech-3.3-8B is more robust to the substantial intra- and inter-speaker variability present in MyST. On SEAME, steering provides limited benefit for Granite-Speech-3.3-8B. This may stem from differences in the model’s pretraining objectives. Although Granite-Speech-3.3-8B supports English-to-Mandarin translation, the speech encoder was not explicitly optimized for Mandarin ASR, potentially resulting in weaker Mandarin acoustic representations for code-switched recognition. SALSA remains close to the zero-shot baseline on both dev-man and dev-sge, while TICL substantially degrades performance, increasing MER by more than 200\% relative to zero-shot. These results suggest that although steering transfers effectively to children's speech, adaptation to code-switching is more challenging and may depend strongly on the pre-training of the underlying model components.

\begin{table}[!ht]
\centering
\small

\begin{tabular}{lcccc}
\toprule
Dataset & $n$ & Zero-shot & TICL & SALSA \\
\midrule

cv-tw
& 200
& 108.36
& 106.32$_{\pm0.00}$
& \textbf{87.54$_{\pm2.10}$} \\

\midrule

\multirow{3}{*}{cv-ru}
& 200
& 87.50
& $122.07_{\pm3.21}$
& $\mathbf{28.33_{\pm0.49}}$ \\

& 500
& 87.50
& $121.90_{\pm3.79}$
& $\mathbf{26.84_{\pm0.34}}$ \\

& 2000
& 87.50
& $119.29_{\pm1.28}$
& $\mathbf{24.65_{\pm0.07}}$ \\

\bottomrule
\end{tabular}
\caption{WER (\%)($\downarrow$) on CommonVoice datasets. SALSA mean $\pm$ std over 5 seeds. Best system per row in \textbf{bold}.}
\label{tab:cv}
\end{table}

\section{Analysis}
To better understand the behavior of learned steering interventions, we analyze the effects of training set size, steering location within the encoder, and the choice of model module being steered.
Prior work on speech representation learning has shown that encoder layers capture progressively different levels of abstraction, ranging from low-level acoustic information in earlier layers to higher-level phonetic and linguistic information in later layers \citep{langedijk2024decoderlens, a-shams-etal-2024-uncovering}. We therefore investigate whether steering effectiveness depends on where interventions are applied within the encoder.

\subsection{Effect of Training Set Size}

Figure~\ref{fig:qwen_icv_scaling} shows the scaling trends for SALSA across varying training set sizes on Qwen2-Audio-7B-Instruct. Across OGI, RSR, and SEAME, increasing the number of steering examples consistently improves performance, with the largest gains typically occurring between 32 and 500 training examples. Beyond approximately 2000 examples, improvements begin to saturate, suggesting diminishing returns from additional steering data. The strongest scaling behavior is observed on RSR and SEAME. On RSR, WER decreases steadily as additional steering examples are introduced, indicating that learned steering vectors can effectively adapt encoder representations to children’s speech despite the substantial acoustic mismatch between adult and child speech. Similarly, both SEAME splits show large improvements with increasing training data, suggesting that steering can effectively adapt multilingual acoustic representations for code-switched ASR without modifying the underlying model parameters.

As noted before, MyST exhibits substantially different behavior. While small amounts of steering data initially improve performance, WER increases at larger training sizes. This suggests that steering may be more sensitive to highly heterogeneous speech distributions containing substantial speaker and acoustic variability. Because steering vectors are shared across all utterances, increasing the amount of highly diverse adaptation data may produce less coherent representation shifts.

Overall, these results suggest that encoder steering is highly data-efficient, with substantial improvements achievable using only a few hundred adaptation examples. This is particularly important for low-resource and out-of-domain speech settings, where collecting large supervised corpora may be impractical.

\subsection{Module-Level Steering Analysis}

\begin{figure}[t]
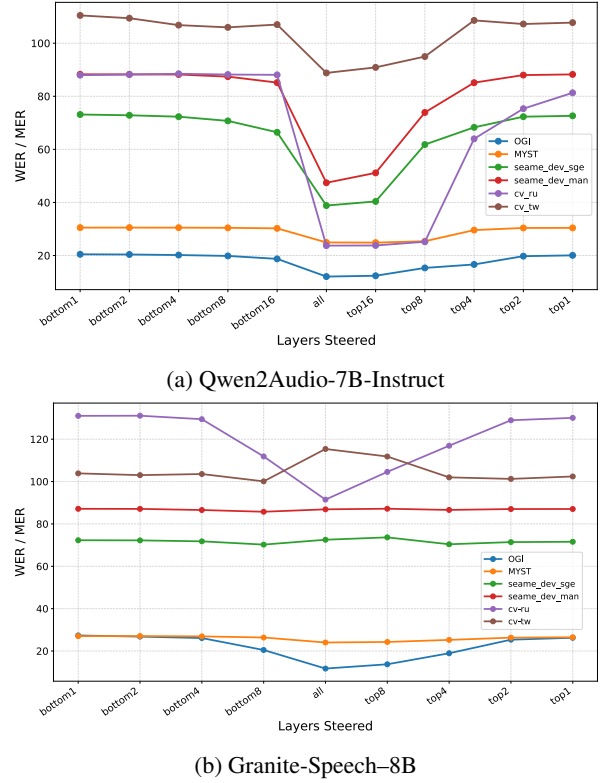

    \centering

    \begin{subfigure}{0.49\textwidth}
        \centering
        \includesvg[width=\linewidth]{plots/qwen_layer_ablation_lineplot}
        \caption{Qwen2Audio-7B-Instruct }
    \end{subfigure}
    \hfill
    \begin{subfigure}{0.49\textwidth}
        \centering
        \includesvg[width=\linewidth]{plots/granite_layer_ablation_lineplot}
        \caption{Granite-Speech--8B}
    \end{subfigure}

    \caption{Performance as a function of encoder steering location. Steering higher encoder layers consistently outperforms steering lower layers across most datasets and models.}
    \label{fig:layers}
\end{figure}

\begin{figure}[t]
    \centering
    \includesvg[width=\linewidth]{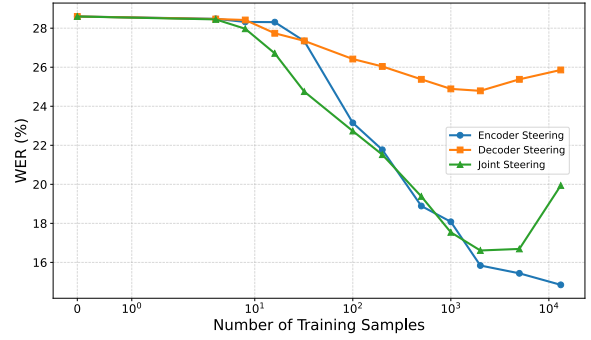}
    \caption{Effect of steering module location on RSR for Qwen2-Audio-7B-Instruct. }
    \label{fig:qwen_icv_modules}
\end{figure}

Figure~\ref{fig:qwen_icv_modules} compares steering applied to the speech encoder, the LLM backbone, and both modules jointly on RSR using Qwen2-Audio-7B-Instruct. Across nearly all training set sizes, encoder steering substantially outperforms decoder-only steering. Decoder steering exhibits only modest improvements over the zero-shot baseline and plateaus quickly as additional training data is introduced. In contrast, encoder steering continues to improve as more steering examples are added, ultimately achieving the best overall performance.

Interestingly, jointly steering both the encoder and decoder does not outperform encoder-only steering. Although joint steering initially matches encoder-only performance at smaller training sizes, the two begin to diverge as additional steering data is introduced. This suggests that modifying decoder representations may partially interfere with the pretrained linguistic representations of the language model backbone. These results suggest that, when both the encoder and language model backbone have prior exposure to the target languages during pretraining, adaptation is more effectively achieved by modifying acoustic representations before projection into the language model rather than intervening directly within the decoder.

\subsection{Layer-Level Steering Analysis}

Figure~\ref{fig:layers} compares performance when steering different layers of the encoder. We evaluate steering applied to progressively larger subsets of lower and upper encoder layers.

For Qwen2-Audio-7B-Instruct, steering higher encoder layers consistently produces larger improvements than steering lower layers alone. On both RSR and OGI, performance approaches that of full encoder steering as progressively more upper encoder layers are steered, whereas steering only the earliest layers provides limited benefit. Similar trends are observed on SEAME and Russian. These results suggest that later encoder representations contain more task-relevant phonetic and linguistic information for adaptation.

Granite-Speech-3.3-8B exhibits a similar but less pronounced pattern. Steering higher encoder layers again provides the strongest improvements on OGI and RSR, although performance varies less dramatically across steering locations compared to Qwen2-Audio-7B-Instruct. This difference may stem from variations in encoder pretraining objectives and architectural design between the two models. However, Twi shows the opposite trend, with lower-layer steering outperforming upper-layer steering. Since Twi is absent from the pretraining data of both the encoder and the language model backbone, this may indicate that adaptation relies more heavily on modifying lower-level acoustic representations when higher-level linguistic representations are not already well established.

More broadly, the degradation observed when steering only lower encoder layers suggests that low-level acoustic representations may already be sufficiently aligned in many settings, while the primary adaptation benefits arise from modifying higher-level phonetic and linguistic representations closer to the projection interface with the language model. Taken together, these results support the hypothesis that encoder steering primarily operates by improving alignment between higher-level speech representations and the pretrained language model decoding space, rather than by substantially altering low-level acoustic processing. These findings are also consistent with prior observations from speech activation steering \citep{sun2026activation}.

\newpage
\section{Conclusion}

We introduced SALSA, a lightweight steering-based adaptation method for SALLMs that directly learns layer-wise steering vectors without requiring paired contrastive examples or modifying backbone model weights. 

Across children’s speech, multilingual speech, and Mandarin-English code-switched speech benchmarks, SALSA substantially improves performance over zero-shot inference and speech in-context learning baselines, demonstrating that simple learned additive interventions are highly effective for out-of-domain ASR settings. By analyzing steering location, encoder depth, and training scale, we further show that adaptation is driven primarily by higher-level encoder representations, whereas steering the language model backbone provides comparatively limited benefit. These findings suggest that adaptation in SALLMs primarily operates through modifying acoustic and phonetic representations before projection into the language model, rather than through interventions within the decoder itself. Our results demonstrate that lightweight representation-level interventions provide an effective and scalable mechanism for adapting pretrained SALLMs to low-resource and out-of-domain speech conditions. Future work may explore input-dependent steering methods that dynamically adapt steering interventions based on the acoustic characteristics of individual utterances, as well as extending steering-based adaptation to other speech settings.

\section{Limitations}

The scope of this work is limited to automatic speech recognition using two speech-aware large language models. The effectiveness and location of steering interventions may vary across different architectures and training objectives. Since SALSA learns a single shared set of steering vectors for all utterances within a dataset, highly heterogeneous speech distributions may therefore require more input-dependent or speaker-specific interventions. Our experiments focus exclusively on automatic speech recognition tasks and a limited set of languages and speech conditions. Broader evaluation across additional tasks, languages, and speaker populations is needed to better understand the generalization and fairness implications of representation-level adaptation methods.

\section{Ethical Considerations}

This work studies adaptation methods for speech-aware large language models in multilingual and out-of-domain speech settings. Although SALSA improves performance for several underrepresented speech conditions, speech recognition systems remain susceptible to demographic and linguistic biases. Steering is substantially more effective for languages already represented during pretraining than for completely unseen languages, highlighting the risk that adaptation methods may reinforce existing imbalances in multilingual speech resources. All pretrained models and datasets used in this work were used in accordance with their respective licenses, access agreements, and terms of use. Some datasets used in this work contain annotations or transcriptions with restricted redistribution terms.



\newpage
\bibliography{custom}

\appendix

\section{Compute and Infrastructure.}
Experiments were conducted using NVIDIA A40 GPUs. Qwen2-Audio-7B-Instruct and Granite-Speech-3.3-8B contain approximately 7B and 8B parameters, respectively. Because SALSA trains only lightweight steering vectors while keeping all backbone parameters frozen, training remained computationally efficient relative to full model fine-tuning.

\section{Dataset Statistics}

\begin{table}[H]
\centering
\small
\begin{tabular}{lccc}
\toprule
Dataset & Train & Dev & Test \\
\midrule
MyST & 73425 & 11732 & 12744 \\
OGI & 50439 & 5482 & 16078 \\
RSR & 13222 & 1467 &  2027 \\
SEAME dev-man& 48297 & 4763 & 0 \\
SEAME dev-sge& 48297 & 1749 & 0 \\
cv-ru & 26920 & 10282 & 10283 \\
cv-tw & 213 & 0 & 30 \\
\bottomrule
\end{tabular}
\caption{Dataset split statistics. Values denote the number of utterances in each split.}
\label{tab:data_stats}
\end{table}

\section{AI Usage}
AI-based assistants were used for limited editing, wording suggestions, and debugging assistance during experiment development and manuscript preparation. All technical content, experimental design, analysis, and conclusions were verified and finalized by the authors.

\clearpage
\onecolumn

\section{Supplementary Tables}
\label{sec:appendix}

\begin{table}[H]
\centering
\small
\label{tab:qwen_icv}
\resizebox{\textwidth}{!}{
\begin{tabular}{lcccccc}
\toprule
$n_{\text{train}}$ & OGI (WER) & MyST (WER) & SEAME Test (MER) & SEAME Val (MER) & CV-RU (WER) & RSR (WER) \\
\midrule
32   & 19.43$\pm$0.72 & 27.27$\pm$0.44 & 67.93$\pm$1.32 & 82.28$\pm$1.08 & 34.78$\pm$1.70 & 27.34$\pm$0.88 \\
100  & 17.09$\pm$0.94 & 25.28$\pm$0.32 & 57.17$\pm$1.77 & 67.37$\pm$1.83 & 29.76$\pm$0.78 & 23.15$\pm$0.53 \\
200  & 15.53$\pm$0.94 & 24.82$\pm$0.67 & 50.41$\pm$1.45 & 60.00$\pm$1.74 & 28.33$\pm$0.49 & 21.77$\pm$0.58 \\
500  & 14.69$\pm$1.32 & 24.92$\pm$0.40 & 44.10$\pm$2.30 & 52.87$\pm$1.61 & 26.84$\pm$0.34 & 18.89$\pm$0.24 \\
1000 & 12.07$\pm$0.45 & 29.15$\pm$1.89 & 41.93$\pm$0.50 & 49.39$\pm$0.87 & 25.83$\pm$0.39 & 18.08$\pm$0.35 \\
2000 & 12.24$\pm$0.97 & 38.68$\pm$8.24 & 40.12$\pm$1.42 & 47.03$\pm$1.67 & 24.69$\pm$0.15 & 15.84$\pm$0.33 \\
5000 & 14.47$\pm$2.68 & 59.01$\pm$15.08 & 41.38$\pm$0.24 & 47.41$\pm$0.85 & 23.72$\pm$0.17 & 15.44$\pm$0.42 \\
\bottomrule
\end{tabular}
}
\caption{Full numerical values for the results shown in Figure~\ref{fig:qwen_icv_scaling}.}
\end{table}

\begin{table}[H]
\centering
\small

\label{tab:qwen_layer_ablation}
\resizebox{\textwidth}{!}{
\begin{tabular}{lcccccccc}
\toprule
Mask & $|S|$ & OGI (WER) & MyST (WER) & SEAME Test (MER) & SEAME Val (MER) & CV-RU (WER) & CV-TW (WER) \\
\midrule
all      & 32 & 12.07$\pm$0.43 & 24.93$\pm$0.40 & 38.83$\pm$1.28 & 47.41$\pm$0.85 & 23.73$\pm$0.19 & 88.80$\pm$1.57 \\
bottom1  & 1  & 20.46$\pm$0.04 & 30.51$\pm$0.02 & 73.13$\pm$0.12 & 88.28$\pm$0.15 & 87.96$\pm$0.06 & 110.47$\pm$1.72 \\
bottom2  & 2  & 20.39$\pm$0.05 & 30.52$\pm$0.02 & 72.84$\pm$0.23 & 88.26$\pm$0.16 & 88.18$\pm$0.30 & 109.42$\pm$1.40 \\
bottom4  & 4  & 20.18$\pm$0.09 & 30.50$\pm$0.03 & 72.31$\pm$0.40 & 88.20$\pm$0.05 & 88.48$\pm$0.31 & 106.81$\pm$2.71 \\
bottom8  & 8  & 19.85$\pm$0.17 & 30.44$\pm$0.03 & 70.73$\pm$0.42 & 87.41$\pm$0.34 & 88.19$\pm$0.36 & 105.97$\pm$2.74 \\
bottom16 & 16 & 18.74$\pm$0.21 & 30.24$\pm$0.04 & 66.44$\pm$0.48 & 85.16$\pm$0.93 & 88.08$\pm$0.41 & 107.02$\pm$2.53 \\
top1     & 1  & 20.06$\pm$0.09 & 30.41$\pm$0.01 & 72.64$\pm$0.11 & 88.24$\pm$0.05 & 81.31$\pm$1.16 & 107.75$\pm$0.96 \\
top2     & 2  & 19.76$\pm$0.16 & 30.39$\pm$0.01 & 72.31$\pm$0.14 & 87.99$\pm$0.05 & 75.33$\pm$2.42 & 107.23$\pm$0.84 \\
top4     & 4  & 16.63$\pm$0.35 & 29.58$\pm$0.15 & 68.29$\pm$0.42 & 85.13$\pm$0.29 & 63.98$\pm$5.05 & 108.59$\pm$7.27 \\
top8     & 8  & 15.33$\pm$0.56 & 25.38$\pm$0.17 & 61.81$\pm$1.40 & 73.90$\pm$4.50 & 25.18$\pm$0.36 & 94.97$\pm$1.89 \\
top16    & 16 & 12.38$\pm$0.48 & 24.86$\pm$0.23 & 40.38$\pm$1.52 & 51.14$\pm$0.48 & 23.79$\pm$0.18 & 90.89$\pm$2.21 \\
\bottomrule
\end{tabular}
}
\caption{Full numerical values for the Qwen2Audio-7B-Instruct results shown in Figure ~\ref{fig:layers}}
\end{table}

\begin{table}[H]
\centering
\small
\label{tab:granite_layer_ablation}
\resizebox{\textwidth}{!}{
\begin{tabular}{lcccccccc}
\toprule
Mask & $|S|$ & OGI (WER) & MyST (WER) & SEAME Test (MER) & SEAME Val (MER) & CV-RU (WER) & CV-TW (WER) \\
\midrule
all      & 16 & 11.72$\pm$0.31 & 24.02$\pm$0.06 & 72.54$\pm$0.71 & 86.89$\pm$0.24 & 91.48$\pm$0.42 & 115.39$\pm$6.24 \\
bottom1  & 1  & 27.36$\pm$0.37 & 27.03$\pm$0.05 & 72.30$\pm$0.06 & 87.14$\pm$0.04 & 131.02$\pm$0.21 & 103.87$\pm$3.53 \\
bottom2  & 2  & 26.79$\pm$0.88 & 27.06$\pm$0.09 & 72.24$\pm$0.10 & 87.10$\pm$0.09 & 131.09$\pm$0.24 & 103.04$\pm$3.65 \\
bottom4  & 4  & 26.12$\pm$0.80 & 26.91$\pm$0.06 & 71.78$\pm$0.25 & 86.57$\pm$0.19 & 129.45$\pm$0.49 & 103.56$\pm$4.31 \\
bottom8  & 8  & 20.48$\pm$0.82 & 26.38$\pm$0.08 & 70.24$\pm$0.41 & 85.75$\pm$0.34 & 111.87$\pm$1.12 & 100.10$\pm$2.42 \\
top1     & 1  & 26.25$\pm$0.16 & 26.49$\pm$0.10 & 71.58$\pm$0.17 & 87.05$\pm$0.03 & 130.08$\pm$0.61 & 102.41$\pm$1.08 \\
top2     & 2  & 25.35$\pm$0.24 & 26.31$\pm$0.10 & 71.42$\pm$0.12 & 87.03$\pm$0.03 & 128.95$\pm$0.52 & 101.26$\pm$0.26 \\
top4     & 4  & 18.94$\pm$0.49 & 25.25$\pm$0.09 & 70.40$\pm$0.50 & 86.61$\pm$0.16 & 116.91$\pm$1.01 & 101.99$\pm$1.88 \\
top8     & 8  & 13.74$\pm$0.37 & 24.27$\pm$0.12 & 73.66$\pm$1.21 & 87.18$\pm$0.46 & 104.57$\pm$2.93 & 111.83$\pm$4.45 \\
\bottomrule
\end{tabular}
}
\caption{Full numerical values for the Granite-Speech-3.3 results shown in Figure ~\ref{fig:layers}}
\end{table}

\begin{table}[H]
\centering
\small
\begin{tabular}{lccc}
\toprule
Num Train Samples & Encoder Steering & Decoder Steering & Joint Steering \\
\midrule
4     & 28.48$\pm$0.31 & 28.47$\pm$0.06 & 28.45$\pm$0.14 \\
8     & 28.32$\pm$0.19 & 28.42$\pm$0.06 & 27.97$\pm$0.09 \\
16    & 28.31$\pm$0.38 & 27.74$\pm$0.52 & 26.71$\pm$0.37 \\
32    & 27.34$\pm$0.88 & 27.35$\pm$0.25 & 24.75$\pm$0.84 \\
100   & 23.15$\pm$0.53 & 26.42$\pm$0.49 & 22.73$\pm$0.47 \\
200   & 21.77$\pm$0.58 & 26.04$\pm$0.34 & 21.52$\pm$0.46 \\
500   & 18.89$\pm$0.24 & 25.38$\pm$0.60 & 19.38$\pm$0.27 \\
1000  & 18.08$\pm$0.35 & 24.89$\pm$0.41 & 17.54$\pm$0.56 \\
2000  & 15.84$\pm$0.33 & 24.79          & 16.61$\pm$0.33 \\
5000  & 15.44           & 25.38$\pm$0.99 & 16.69$\pm$0.32 \\
13222 & 14.85           & 25.86           & 19.94 \\
\bottomrule
\end{tabular}
\caption{Full numerical values for the results shown in Figure~\ref{fig:qwen_icv_modules}.}
\label{tab:module_scaling}
\end{table}

\end{document}